\def\year{2021}\relax
\documentclass[letterpaper]{article} 
\usepackage{aaai21}  
\usepackage{times}  
\usepackage{helvet} 
\usepackage{courier}  
\usepackage[hyphens]{url}  
\usepackage{graphicx} 

\usepackage{natbib}  
\usepackage{caption} 
\usepackage{latexsym}

\usepackage{soul}
\usepackage{url}
\usepackage{caption}
\usepackage{graphicx}
\usepackage{amssymb}
\usepackage{amsmath}
\usepackage{booktabs}
\usepackage{todonotes}
\usepackage{microtype}
\usepackage{tabularx}
\usepackage{booktabs}
\usepackage{multirow}
\usepackage{algorithm}
\usepackage{algorithmic}
\usepackage[normalem]{ulem}
\usepackage{xcolor}
\usepackage{xr}
\usepackage[switch]{lineno} 

\makeatletter
\newcommand*{\addFileDependency}[1]{
  \typeout{(#1)}
  \@addtofilelist{#1}
  \IfFileExists{#1}{}{\typeout{No file #1.}}
}
\makeatother

\newcommand*{\myexternaldocument}[1]{%
    \externaldocument{#1}%
    \addFileDependency{#1.tex}%
    \addFileDependency{#1.aux}%
}
\myexternaldocument{app}

\newcommand\muc[0] {MUC}
\newcommand\bcubed[0] {$\text{B}^3$ }
\newcommand\ceaf[0] {$\text{CEAF}_{\phi_4}$}
\newcommand{\fone}[0] {F$_1$ }

\title{Joint Semantic Analysis with Document-Level Cross-Task Coherence Rewards}

\author{
    Rahul Aralikatte, Mostafa Abdou, Heather Lent,\\ Daniel Hershcovich and Anders S{\o}gaard\\
}
\affiliations{
    University of Copenhagen\\
    \{rahul, abdou, hcl, dh, soegaard\}@di.ku.dk
}

\begin{document}
\maketitle

\begin{abstract}
Coreference resolution and semantic role labeling are NLP tasks that capture different aspects of semantics, indicating respectively, which expressions refer to the same entity, and what semantic roles expressions serve in the sentence. However, they are often closely interdependent, and both generally necessitate natural language understanding. Do they form a coherent abstract representation of documents? We present a neural network architecture for joint coreference resolution and semantic role labeling for English, and train graph neural networks to model the \textit{coherence} of the combined shallow semantic graph. Using the resulting coherence score as a reward for our joint semantic analyzer, we use reinforcement learning to encourage global coherence over the document and between semantic annotations. This leads to improvements on both tasks in multiple datasets from different domains, and across a range of encoders of different expressivity, calling, we believe, for a more holistic approach to semantics in NLP.
\end{abstract}

\section{Introduction}\label{sec:intro}

\begin{figure}[t]
    \centering
    \includegraphics[width=\columnwidth]{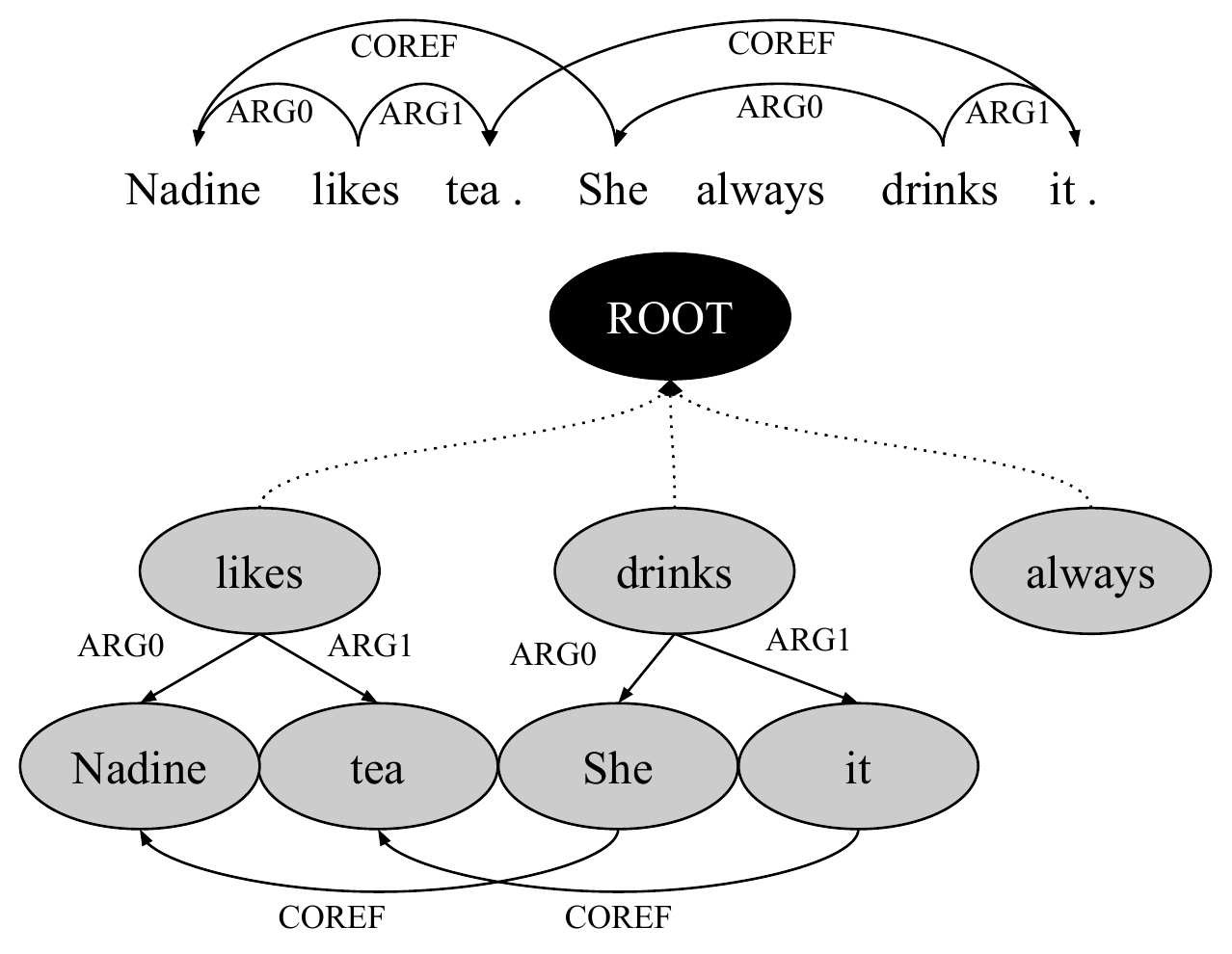}
    \caption{Example coreference and semantic role annotation for a two-sentence document. Top: the original annotation shown as dependencies. Bottom: shallow semantic graph (SSG), where sub-graph heads are connected (with dotted lines) to a dummy root node.}
    \label{fig:example}
\end{figure}

Coreference resolution and semantic role labeling (SRL)
contribute in complimentary ways to forming coherent discourse representations. SRL establishes predicate-argument relations between expressions, and coreference resolution determines what entities these expressions refer to. While often
treated separately \cite{he-etal-2017-deep,he-etal-2018-jointly,lee-etal-2017-end,lee-etal-2018-higher, Joshi_2019},
some frameworks consider coreference and semantic roles part of a more holistic meaning representation \cite{shibata-kurohashi-2018-entity}.
For example, the Groningen Meaning Bank \cite{bos2017groningen} annotates 
documents with discourse representation structures \cite{kamp2013discourse}, which subsume both levels of analysis; the same holds for other meaning representation frameworks, such as UCCA \cite{abend-rappoport-2013-universal,prange-etal-2019-semantically} and
AMR \cite{banarescu-etal-2013-abstract,ogorman-etal-2018-amr}.
However, these frameworks do not offer the simplicity of SRL and coreference annotation, and perhaps consequently require more effort to annotate, and do not have the same amounts of training data \cite{abend-rappoport-2017-state}.
Furthermore, comprehensive meaning representation \textit{parsing} approaches \cite{liu-etal-2018-discourse,hershcovich-etal-2017-transition,cai-lam-2020-amr} tend to be more complex than the sequence tagging or span-based models often used for coreference resolution and SRL, often referred to as \textit{shallow semantic parsing}.

In this paper, we investigate a ``minimal'' approach to discourse-level semantic parsing, combining coreference and semantic roles in \textit{shallow semantic graphs} (SSGs) that can be seen as a simple, yet rich, discourse-level meaning representations.
Consider the two sentences shown in Figure~\ref{fig:example}, augmented with a (partial) annotation of coreference and semantic roles.
A coreference resolver is expected to resolve \textit{Nadine} as an antecedent of \textit{she}, and \textit{tea} as an antecedent of \textit{it}, since these mentions refer to the same entities. 
A semantic role labeler is expected to detect that these entities are arguments of the predicates {\em like} and {\em drink}. 
The overall semantic analysis corresponds to a coherent and common situation, where someone likes something and consumes it---a very plausible interpretation. This paper presents a model that scores the plausibility or \textit{coherence} of an interpretation based on merged SRL and coreference graphs, or SSGs. While Figure~\ref{fig:example} is a simple example that existing SRL and coreference systems will likely handle well, we explore whether such systems in general benefit from feedback from a model that rewards the coherence of their output.

\paragraph{Contributions} 
We jointly model coreference resolution and SRL to form discourse-level semantic structures, or SSGs (\S\ref{sec:models}). We explicitly model their coherence, presenting a reinforcement learning architecture for semi-supervised fine-tuning of coreference resolvers and semantic role labelers with coherence rewards on unlabeled data (\S\ref{sec:semisupervised}), 
improving both coreference resolution and SRL.
We present experiments across six encoders of different complexities, six different coreference resolution datasets, and four different SRL datasets (\S\ref{sec:exp}), showing improvements across all encoders for coreference resolution, and on 4/6 for SRL, for single-task setups; and similar improvements in multi-task setups, where encoder parameters are shared across the two tasks (\S\ref{sec:results}).
Finally, we analyze the results (\S\ref{sec:analysis}), showing that our fine-tuning setup is particularly beneficial for smaller documents while being on-par with strong baselines on larger documents and that the majority of the remaining coreference errors occur when the antecedent is a pronoun. 

\section{Joint Coreference Resolution and SRL}\label{sec:models}

\begin{figure}[t]
    \centering
    \includegraphics[width=\columnwidth]{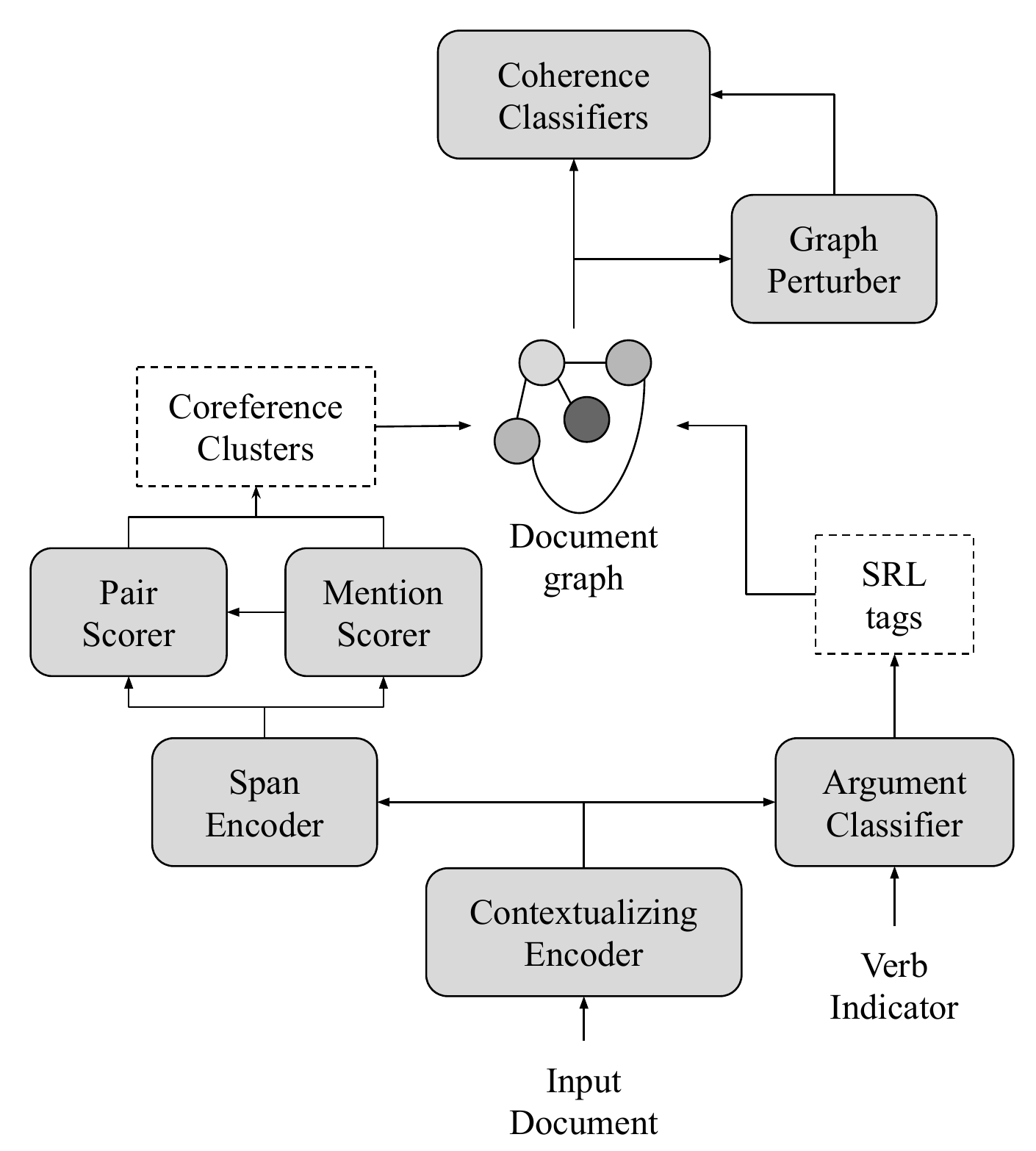}
    \caption{Joint coreference resolution and SRL (bottom half) with a coherence objective (top half). The contextualizing encoder is shared in the multi-task setup, and separate in the single-task one. Predictions from the coreference and SRL models are combined to a document-level SSG, which is scored by coherence classifiers to reward the models.}
    \label{fig:arch}
\end{figure}

We build baseline single-task and multi-task \textit{supervised models} for coreference resolution and SRL. The overall model architecture is illustrated in Figure~\ref{fig:arch} (bottom half; till the coreference clusters and SRL tags are generated). In the multi-task setup only the contextualizing encoder is shared. In the single-task setup no parameters are shared.

\paragraph{Coreference Resolver} The coreference model is based on the architecture presented in \citet{lee-etal-2017-end}. Each token's embedding is obtained using a contextualizing encoder. Using a \textit{span encoder}, the token embeddings are combined into span representations $s(i, j)$, 
where $i$ and $j$ are the start and end indices 
in the document. Each span is represented as the concatenation of: (i) its first and last token embeddings, and (ii) an attention-based aggregation of embeddings of all tokens in the span. These span representations are pruned with a \textit{mention scorer}, which outputs the probability of $s(i, j)$ being a coreferent mention. Next, the mention representations are paired together and scored again with a \textit{pair scorer}, which predicts the probability of the mentions referring to each other. Coreferring mentions are collected to form clusters. This architecture is combined with pre-trained language models in \citet{lee-etal-2018-higher} and \citet{Joshi_2019} to get state-of-the-art results.

\paragraph{Semantic Role Labeler} The SRL tagger is based on the architecture presented in \citet{he-etal-2017-deep}. The model uses the contextualizing encoder to embed tokens which are concatenated with a binary indicator to identify whether the token is a verb or not. These token representations are presented to a \textit{argument classifier} for BIO sequence tagging. The current state-of-the-art \cite{he-etal-2018-jointly} uses an architecture similar to that of \citet{lee-etal-2017-end}, where it jointly predicts both arguments and predicates.

\paragraph{Contextualizing Encoder}
In all setups, we experiment with (i) an LSTM + CNN encoder, and (ii) five {\sc BERT} \cite{devlin-etal-2019-bert} encoders of different sizes.
In the LSTM + CNN encoder, a bi-LSTM contextualizes words embedded with GloVe \cite{pennington-etal-2014-glove} and a CNN encodes individual characters. The final representation is the concatenation of the two.
For the {\sc BERT} encoders, we experiment with different encoder sizes as shown in Table \ref{tab:bert}, using each token's wordpiece embeddings.
Encoder hyperparameters are given in \S\ref{sec:implementation}.

\section{Semi-Supervised Fine-Tuning}\label{sec:semisupervised}


In the semi-supervised stage of training, classifiers trained on SSGs created from labeled data (Figure~\ref{fig:example}) are used to fine-tune the supervised models on unlabeled data by reinforcement learning. For each unlabeled document, we use the predicted annotations of the supervised models to build an SSG consisting of SRL predicates and arguments, with links between coreferent mentions. Edge labels are used to distinguish between SRL and coreference edges. These graphs are scored by \textit{graph classifiers} (\S\ref{sec:classifiers}), trained using graph perturbations (\S\ref{sec:perturb}) to model semantic coherence. The confidence value is used as a reward to fine-tune the supervised models using policy gradient (\S\ref{sec:fine-tune}).

\subsection{Coherence Classifiers}\label{sec:classifiers}
We use a graph convolution network \cite[GCN;][]{DBLP:conf/iclr/KipfW17} to construct continuous representations of the SSGs, where a node representation is composed via a learnt weighted sum of neighboring nodes. Since nodes correspond to text spans, to initialize their representations, we use the supervised model's span encoder.
To get the final graph encoding, all the node representations are averaged and compressed using the logistic function as shown in Equation~\ref{eqn:graph-rep}.
\begin{equation}
    \text{graph}_{enc} = \sigma\left(\frac{1}{N} \sum_{i=1}^{N} \text{node}_{enc}^{i}\right) \label{eqn:graph-rep}
\end{equation}
The GCN parameters are pre-trained using deep graph infomax \cite[DGI;][]{velivckovic2018deep}, which relies on graph perturbations to learn a task-independent representation.
We contrastively train the GCN encoder on gold and \textit{perturbed} graphs, which are generated by randomly perturbing the gold graphs (\S\ref{sec:perturb}). We then use the same perturbations to train a logistic regression classifier, with the GCN outputs as features, to discriminate gold graphs from perturbed graphs.
As shown in \S\ref{sec:coherence_results}, the trained classifiers are almost perfectly accurate on an unseen development set.

The process for training the coherence classifiers is shown in Algorithm~\ref{alg:cc}. First an SSG $g \in \mathbb{G}$ is built for each labeled document. Then for each type of perturbation $p \in P$, we train one classifier as follows: (i) perturb $g$ to get $g_{p}$ using perturbation $p$. We use a decay factor $d \in \{0, 1\}$ to decide the probability of perturbing a sentence in the document. We start with $d=0.8$ and decay it till $d=0.1$, (ii) once we have a list of perturbed graphs $\mathbb{G}_p$, we train the GCN using DGI, which uses a contrastive loss to learn graph representations such that each pair ($g, g_{p}$) is as different to each other as possible, (iii) we use the GCN to get the final representations of graphs in $\mathbb{G}$ and $\mathbb{G}_p$ and create a training dataset consisting of the following (graph, label) pairs: $\{(g, 1) : g\in\mathbb{G}\} \cup \{(g_{p} , 0) : g_{p}\in\mathbb{G}_p\}$, and (iv) we train a logistic regression classifier.

\begin{algorithm}[t]
\caption{\label{alg:cc} Training Coherence Classifiers}
\begin{algorithmic}
\REQUIRE $\mathbb{G}$: List of SSGs \\
\REQUIRE P: List of perturbations to perform \\
\REQUIRE d: Decay factor \\
\STATE Initialize clfs = $\emptyset$ \\
\FOR{p in P}
    \FOR{epoch = 1, \ldots, N}
        \STATE Initialize $\mathbb{G}_p$ = $\emptyset$ \\
        \FOR{g in $\mathbb{G}$}
            \STATE g$_{p}$ = p (g, d)
            \STATE $\mathbb{G}_p$.add (g$_{p}$) \\
        \ENDFOR
        \STATE encoder = DGI ($\mathbb{G}, \mathbb{G}_p$) \\
        \STATE d = decay (d) \\
    \ENDFOR
    \STATE data$_{+}$ = (encoder ($\mathbb{G}$), 1)
    \STATE data$_{-}$ = (encoder ($\mathbb{G}_p$), 0)
    \STATE clf$_{p}$ = logistic (data$_{+}$, data$_{-}$) \\
    \STATE clfs.add (clf$_{p}$) \\
\ENDFOR
\RETURN clfs
\end{algorithmic}
\end{algorithm}

\subsection{Graph Perturbations}\label{sec:perturb}

\begin{figure*}[t]
    \centering
    \includegraphics[width=\textwidth, height=1.5in]{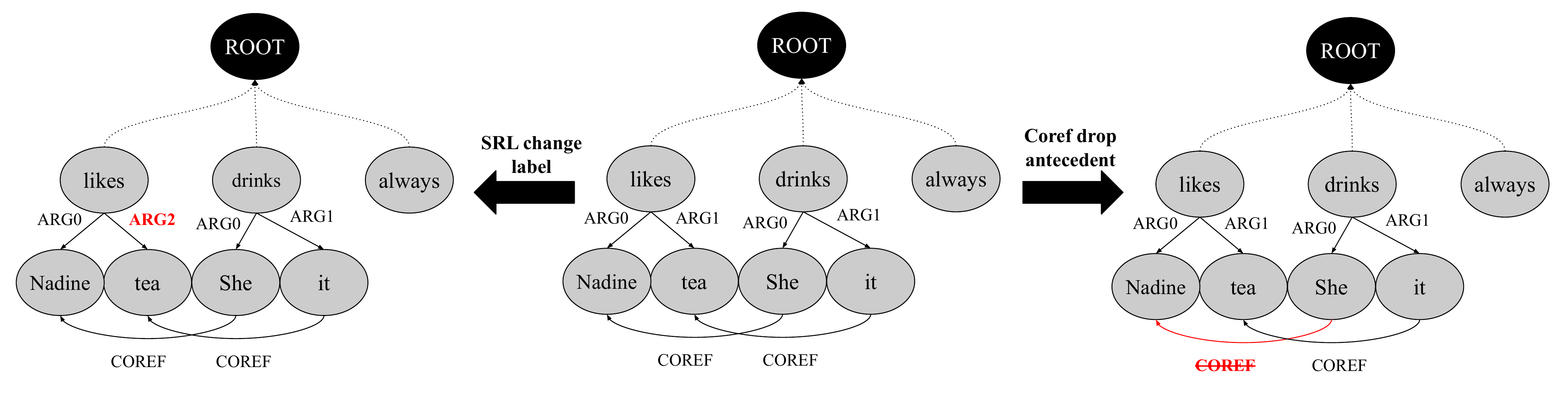}
    \caption{Examples for graph perturbations, starting from the SSG in Figure~\ref{fig:example} (center). An `SRL change label' perturbation is applied to generate a graph (left), where {\sc ARG1} is changed to {\sc ARG2}. A `Coref drop antecedent' perturbation is applied to generate a graph (right) where a {\sc coref} edge is deleted.}
    \label{fig:perturb}
\end{figure*}

To train the GCN with DGI, we perturb the gold graphs to reflect the statistics of errors made by the supervised models we want to fine-tune. In general, perturbations are sampled from the following operations:
(i) randomly removing edges,
(ii) randomly adding edges between existing nodes with a random label, or
(iii) randomly adding nodes with a span that is a constituent in the sentence, and a random edge to another existing node.
We arbitrarily choose to sample SRL and coreference perturbations with a 3-to-1 ratio.  

For SRL perturbations,
we rely on the error analysis made by \citet{he-etal-2017-deep}, whose SRL model is the basis for ours: 29.3\% of errors correspond to incorrect argument labels; 4.5\% to moved unique arguments; 10.6\% to split arguments; 14.7\% to merged arguments; 18\% to incorrect boundaries; 7.4\% to superfluous arguments; and 11\% to missed arguments. Consequently, we sample perturbations proportionally to the corresponding error's frequency. We further use \citet{he-etal-2017-deep}'s observed confusion matrix of predicted and gold argument labels, sampling replacement labels accordingly.
For \textit{coreference} perturbations, we add a random edge between existing nodes or remove an edge, with uniform probability.

We train one classifier to identify each type of perturbation, resulting in nine different classifiers (seven for SRL and two for coreference; an example for one of each is illustrated in Figure~\ref{fig:perturb}). The final confidence for a graph is the average of the individual classifier confidence scores.

\subsection{Model Fine-Tuning}\label{sec:fine-tune}

Finally, we use the learned classifiers to fine-tune the underlying coreference resolver and semantic role labeler;
using plain text from summary paragraphs of Wikipedia articles, we apply the supervised models to sample an SSG. Using the coherence classifiers' confidence score as a reward, we train the models with policy gradient.

During policy gradient, we consider the selection of SSG edges as actions. More concretely, for coreference resolution, picking the antecedent to each mention is considered an action. Therefore from Figure \ref{fig:example}, assuming the model found four mentions (`Nadine', `tea', `She', and `it'), it takes four actions (connecting `Nadine$\rightarrow\phi$', `tea$\rightarrow\phi$', `she$\rightarrow$Nadine', `it$\rightarrow$tea').\footnote{$\phi$ indicates no antecedent} For SRL, assigning a label to a token is considered as an action. Therefore the model has to perform nine actions (one for each token) to label Figure \ref{fig:example}.

In this work, we assume that all actions are equally good and reward them uniformly. Assigning rewards to individual actions would probably yield better results but is non-trivial and left for future exploration. 

\section{Experiments}\label{sec:exp}
In this section, we briefly describe the datasets used to train and evaluate our models before moving on to the experimental setup. We then provide implementation details for each stage of the training process and finally present the results of our experiments.

\subsection{Datasets}\label{sec:datasets}
For supervised training, we use data from the CoNLL-2012 shared task \cite{pradhan-etal-2012-conll},
which contains data from OntoNotes 5.0\footnote{\url{https://catalog.ldc.upenn.edu/LDC2013T19}}
with annotations for both coreference resolution and semantic role labeling.

As additional out-of-domain (OOD) development and test data for coreference resolution, we use
(i) PreCo \cite{preco}, which contains web-crawled documents and data from the RACE dataset \cite{race};
(ii) Phrase Detectives \cite{phrase-detectives}, which contains two evaluation sets, one sampled from Wikipedia and the other from the Gutenberg project;
(iii) WikiCoref \cite{wikicoref}, which contains long form documents from the English Wikipedia; and 
(iv) WinoBias \cite{winobias}, which is focused on gender bias with Winograd-schema style sentences, authored manually.

For SRL, we additionally use
(i) the CoNLL-2005 shared task data \cite{conll05}, which contains two evaluation sets: the in-domain WSJ set and the OOD Brown set; and
(ii) English Web Treebank \cite{ewt}\footnote{\url{https://catalog.ldc.upenn.edu/LDC2017T15}}, which contains weblogs, newsgroups, email, question-answers and review text.

\subsection{Experimental Setup}\label{sec:setup}
We first train the coreference and SRL models (\S\ref{sec:models}) using supervised learning,
and the coherence classifiers on gold graphs and their perturbations.
Both are trained on the CoNLL-2012 training set.
We then fine-tune the models by semi-supervised learning (\S\ref{sec:semisupervised}), with the summary paragraphs of 10,000 randomly sampled English Wikipedia articles.\footnote{\url{https://www.wikipedia.org}, dump from March 4, 2019.}
We test our models across six domains for coreference resolution, and four domains for SRL, using in-domain evaluation data.

\subsection{Implementation Details}\label{sec:implementation}

\begin{table}[t]
    \centering
    \small
    \begin{tabular}{>{\raggedright\arraybackslash}m{3cm}|
                    >{\raggedright\arraybackslash}m{1cm}
                    >{\raggedright\arraybackslash}m{1cm}
                    >{\raggedright\arraybackslash}m{1cm}}
        \toprule
        \textbf{Hyperparameters} & \textbf{Lee et al. (2018)} & \textbf{\citet{Joshi_2019}} & \textbf{Ours} \\
        \midrule
        max. span width & 30 & 30 & 10 \\
        cxt. enc. (layers/dims) & 3/1024 & 24/1024 & 12/768* \\
        span enc. (layers/dims) & 3/400 & - & 1/400 \\
        pruner (layers/dims) & 2/150 & 1/1000 & 1/150 \\
        top span ratio & 0.4 & 0.4 & 0.3 \\
        max antecedents & 250 & 50 & 100 \\
        course to fine inference & True & True & False \\
        \bottomrule
    \end{tabular}
    \caption{Comparison of hyperparameters between state-of-the-art and our coreference models. $^*$This value is for {\sc BERT}-Base. See Table \ref{tab:bert} for other sizes.}
    \label{tab:coref-model-cmp}
\end{table}

Since the goal of this work is not to surpass the state of the art, but to demonstrate that discourse-level coherence can be used to improve shallow semantic analysis, and due to memory and compute constraints, we use smaller versions of the best performing architectures in the literature as baselines.

\paragraph{Coreference model}

We use the same architecture that state-of-the-art coreference systems like \citet{lee-etal-2017-end, lee-etal-2018-higher} and \citet{Joshi_2019} use, but with lesser capacity. A comparison of the important hyperparameters that vary between our model and the current state-of-the-art is shown in Table~\ref{tab:coref-model-cmp}.

\paragraph{SRL model}
\citet{he-etal-2017-deep} use 8 LSTM layers with highway connections and recurrent dropout. We replace this encoder with each of our contextualizing encoder configurations. Following \citet{he-etal-2017-deep}, we also use constrained decoding to produce only valid BIO tags as output.

\paragraph{Contextualizing encoders}
For the LSTM + CNN encoder, 300-dimensional GloVe embeddings \cite{pennington-etal-2014-glove} are fed into a bi-LSTM with a hidden size of 200, to get a 400-dimensional word representation. We concatenate this with 100-dimensional character embeddings obtained from a CNN character encoder with a filter size of 5. The other five encoders are based on the standard {\sc BERT} recipe \cite{bert-sizes}, and their sizes can be seen in Table \ref{tab:bert}.

\begin{table}[t]
    \centering
    \begin{tabular}{lcc}
        \toprule
        \textbf{Encoder} & \textbf{\# layers} & \textbf{dim}  \\
        \midrule
        LSTM + CNN & 1 & 500 \\
        {\sc BERT}-Tiny & 2 & 128 \\
        {\sc BERT}-Mini & 4 & 256 \\
        {\sc BERT}-Small & 4 & 512 \\
        {\sc BERT}-Medium & 8 & 512 \\
        {\sc BERT}-Base & 12 & 768 \\
        \bottomrule
    \end{tabular}
    \caption{Number of layers and the output dimension of our contextualizing encoders.}
    \label{tab:bert}
\end{table}

\paragraph{Supervised training} For training both single-task and multi-task models, we use the Adam optimizer \cite{adam} with a weight decay of $0.01$ and initial learning rate of $10^{-3}$. For {\sc BERT} parameters, the learning rate is lowered to $10^{-5}$. We reduce the learning rates by a factor of $2$ if the evaluation on the development sets does not improve after every other epoch. The training is stopped either after $100$ epochs, or when the minimum learning rate of $10^{-7}$ is reached. In the multi-task setup, we sample a batch from each task with a frequency proportional to the dataset size of that task. All experiments are run on a single GPU with $16$GB memory. The hyperparameters were manually selected to accommodate for training time and resource limitation, and were not tuned based on model evaluation.

\paragraph{Coherence Classifiers} The GCN encoder used to encode the SSGs has $512$ hidden channels and is trained with Adam for $10$ epochs. We use a 20-dimensional embedding to represent the type of node and a binary indicator to represent the edge type.

\paragraph{Fine-tuning} The supervised models are fine-tuned for 10 epochs with the same optimizer configuration. Only the learning rate is changed to $3\cdot10^{-4}$. Hill climbing is used during policy gradient, i.e., if fine-tuning on a batch of Wikipedia documents does not yield an improvement, the parameters are reset to their previous best state. 

In the multi-task setup, the coreference resolution and SRL sub-models are fine-tuned separately. This is because we do not want to sample actions for both tasks as it makes the constructed SSG more noisy. For constructing the SSGs in the single-task setup, we use the best performing SRL model for fine-tuning the coreference resolution model, and the best performing coreference resolution model for fine-tuning the SRL model.

\section{Results}\label{sec:results}

\subsection{Coreference Resolution and SRL}\label{sec:coref_srl_results}

The mean \fone over \muc, \ceaf, and \bcubed scores averaged across the six test sets for coreference resolution and four test sets for SRL (including in-domain and out-of-domain), for each of the six encoder configurations, is presented in Table~\ref{tab:main-results}. The individual results for each dataset is presented in Tables~\ref{tab:coref-single-task-results} and~\ref{tab:srl-single-task-results} in Appendix~\ref{sec:single-task_results} for single-task models, and in Tables \ref{tab:coref-multi-task-results} and \ref{tab:srl-multi-task-results} for multi-task models respectively.

\begin{table}
    \centering
    \begin{tabular}{@{}ll|c@{}}
        \toprule
        \multicolumn{2}{c|}{\textbf{Perturbation type}} & \textbf{Accuracy (\%)} \\
        \midrule
        \multirow{7}{*}{SRL} & change label & 98.98 \\
        & move argument & 99.88 \\
        & split spans & 99.72 \\
        & merge spans & 99.29 \\
        & change boundary & 98.96 \\
        & add argument & 99.22 \\
        & drop argument & 100.00 \\
        \midrule
        \multirow{2}{*}{Coref} & add antecedent & 99.10 \\
        & drop antecedent & 100.00 \\
        \bottomrule
    \end{tabular}
    \caption{Graph classifier development accuracy.}
    \label{tab:clf-result}
\end{table}

We see substantial improvements from coherence fine-tuning across the board for all coreference tasks. Results for single-task SRL improves in all settings except for {\sc BERT}-mini and {\sc BERT}-medium encoders. In the multi-task setting for SRL, we see consistent improvements with two exceptions: the results for LSTM + CNN and {\sc BERT}-base. Coreference resolution generally improves more than for SRL.

\subsection{Coherence Classifiers}\label{sec:coherence_results}

The accuracy of the nine coherence classifiers (\S\ref{sec:classifiers}) on the CoNLL-2012 development set is shown in Table~\ref{tab:clf-result},
showing that the classifiers can almost perfectly detect perturbed graphs, and explaining their effectiveness at providing a reward signal to the models.
While it could be argued that this indicates that the perturbations are too easy to detect, observing the perturbed graphs (exemplified in Figure~\ref{fig:perturb}) leads to the impression that they require sensitivity to distinctions that are important for correct coreference resolution, SRL and the coherence between them. Indeed, the rewards lead to improvements in each of the tasks.

\begin{table*}[th]
\centering
\begin{tabular}{@{}r|cccc|cccc@{}}
\toprule
& \multicolumn{4}{c|}{\textbf{Single-Task}} & \multicolumn{4}{c}{\textbf{Multi-Task}} \\
& \multicolumn{2}{c}{\textbf{Coreference}} & \multicolumn{2}{c|}{\textbf{SRL}} & \multicolumn{2}{c}{\textbf{Coreference}} & \multicolumn{2}{c}{\textbf{SRL}} \\
\multirow{-3}{*}{\textbf{Encoder}} & {\textbf{Baseline}} & {\textbf{Ours}} & {\textbf{Baseline}} & {\textbf{Ours}} & {\textbf{Baseline}} & {\textbf{Ours}} & {\textbf{Baseline}} & {\textbf{Ours}} \\
\midrule
{\textbf{LSTM + CNN}}         & {49.01} & {\bf 49.40} & {67.63}     & {\bf 67.74} & {48.65} & {\bf 49.60} & {\bf 67.28} & {67.05} \\
{\textbf{{\sc BERT}-Tiny}}    & {49.70} & {\bf 50.95} & {56.87}     & {\bf 57.08} & {45.65} & {\bf 51.17} & {56.65}     & {\bf 56.85} \\
{\textbf{{\sc BERT}-Mini}}    & {52.61} & {\bf 52.88} & {\bf 70.51} & {70.48}     & {50.14} & {\bf 53.02} & {71.10}     & {\bf 71.13} \\
{\textbf{{\sc BERT}-Small}}   & {52.76} & {\bf 53.90} & {74.26}     & {\bf 74.48} & {51.26} & {\bf 53.73} & {74.72}     & {\bf 74.77} \\
{\textbf{{\sc BERT}-Medium}}  & {55.67} & {\bf 56.19} & {\bf 75.62} & {75.57}     & {51.48} & {\bf 55.52} & {77.89}     & {\bf 78.01} \\
{\textbf{{\sc BERT}-Base}}    & {57.78} & {\bf 58.18} & {79.46}     & {\bf 79.52} & {56.40} & {\bf 57.55} & {\bf 80.25} & {80.19} \\
\bottomrule
\end{tabular}
\caption{{\sc Coreference resolution} and {\sc Semantic role labeling} results of single-task and multi-task models. `Baseline' and `Ours' represent the the supervised baseline and coherence fine-tuned models respectively. The numbers are the mean of \muc, \bcubed and \ceaf \fone scores averaged over six (four) coreference (SRL) datasets.}
\label{tab:main-results}
\end{table*}

\section{Error Analysis}\label{sec:analysis}

By analysing the results of the fine-tuned models on all datasets (Table \ref{tab:main-results}), we make the following observations:\footnote{Unless mentioned otherwise, all analysis is carried out on the single-task {\sc BERT}-Base model.}

\paragraph{Document length}

\begin{figure}[t]
    \centering
    \includegraphics[width=\columnwidth]{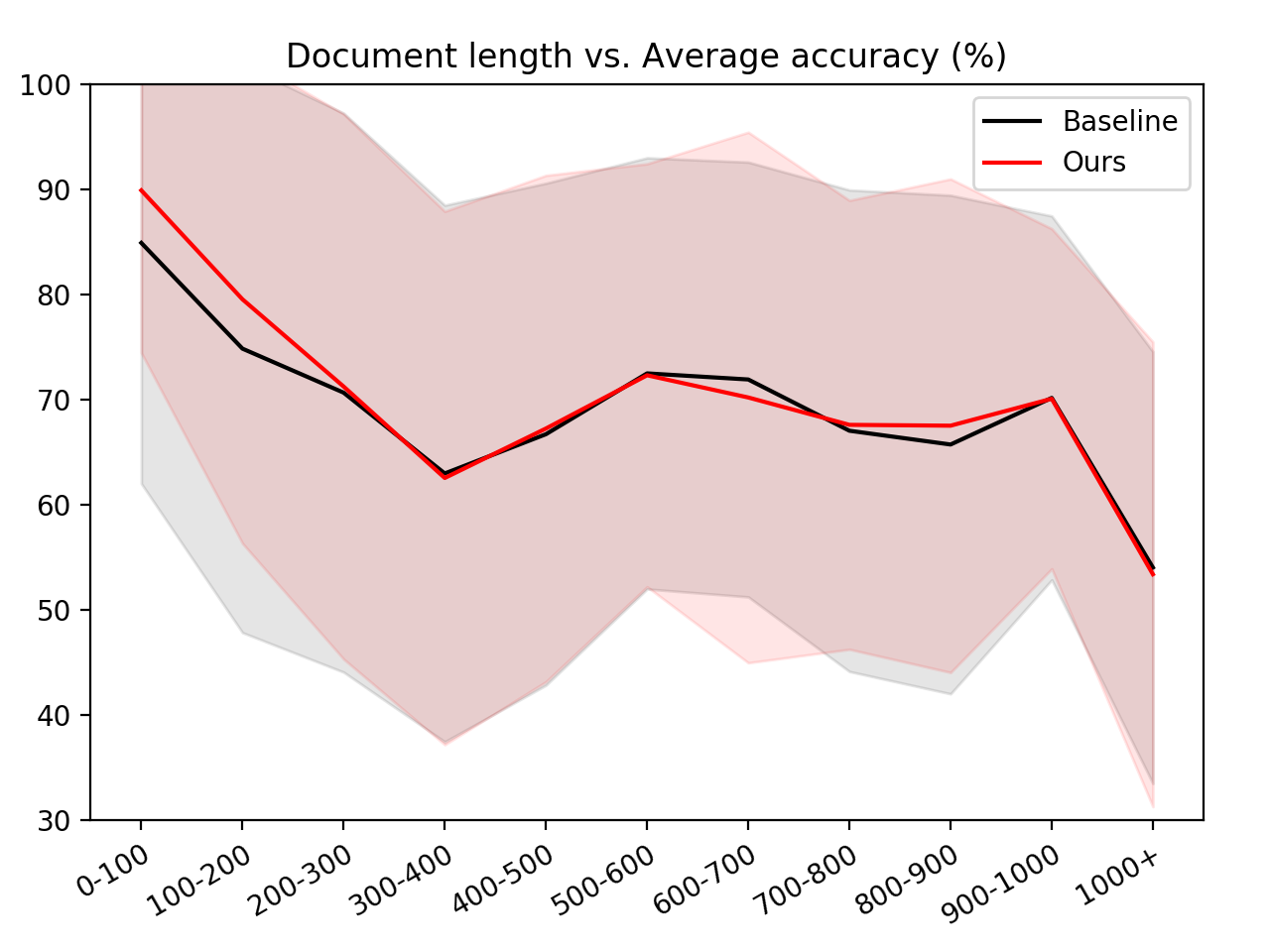}
    \caption{Percentage of correct predictions of our {\sc BERT}-Base coreference model across all datasets plotted against document lengths.}
    \label{fig:doc_errors}
\end{figure}

Fine-tuning leads to larger improvements on smaller documents (see Figure \ref{fig:doc_errors}). This is likely because the unlabeled data we use for fine-tuning consists of short paragraphs. While using longer documents for fine-tuning was not possible due to memory constraints, we expect that this will increase the model's sensitivity to long-distance inter-dependencies, and further improve its performance on these documents.

\paragraph{Coreference resolution vs. SRL}
In general, SRL sees smaller improvements from fine-tuning with policy gradient than coreference resolvers,
probably because it is harder to assign credit to specific model decisions \cite{Langford:Zadrozny:05}. Semantic role labeling of a paragraph typically requires a much longer sequence of actions than determining coreference, leading to limited benefit from reinforcement learning. Similar results have been observed in machine translation \cite{choshen2019weaknesses}.

\paragraph{Precision vs. recall}
Precision often increases after fine-tuning whereas recall decreases. Similar effects have been reported for knowledge-base grounding of coreference resolvers \cite{aralikatte-etal-2019-rewarding}. 

\paragraph{Encoder sizes}
From the results, we also see that our fine-tuning approach is robust to encoder sizes with improvements across the board. It is particularly interesting to see that the multi-task {\sc BERT}-Tiny coreference models come close or even surpass the bigger {\sc BERT}-Base models on datasets like PreCo and WinoBias, which contain short documents (see Table \ref{tab:coref-multi-task-results} in Appendix~\ref{sec:single-task_results}).

In both single-task and multi-task setups, fine-tuning helps the smaller coreference models more than the larger ones, which are already more accurate. This trend is expected as the larger models tend to be over-parameterized.

\begin{figure}[t]
    \centering
    \includegraphics[width=\columnwidth]{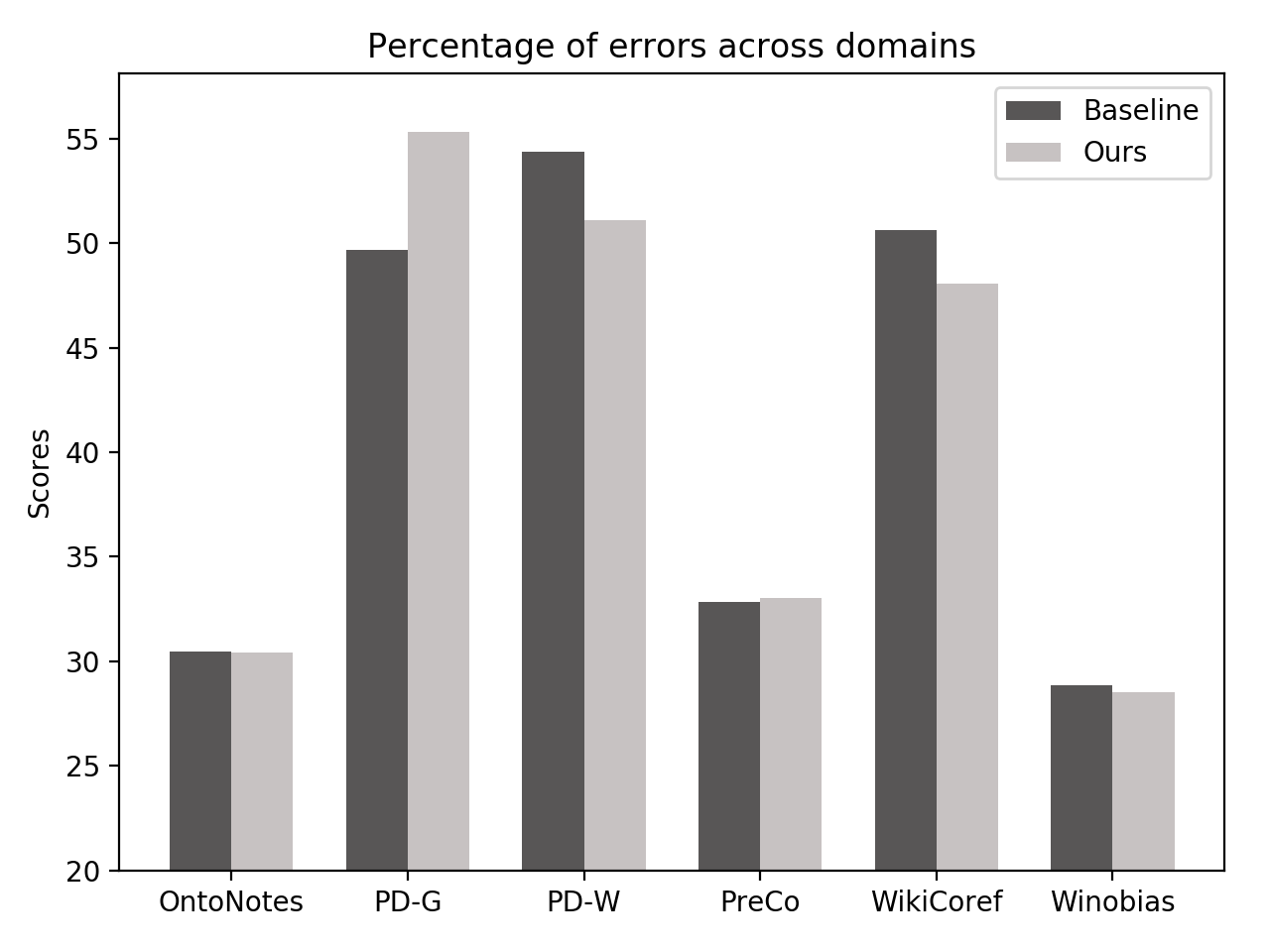}
    \caption{Percentage of errors over the total amount of predictions that our coreference system makes across each domain of the evaluation data.}
    \label{fig:domainerror}
\end{figure}

\paragraph{Domain adaptation}
We also perform an error analysis to identify the domains which are hard for our coreference models (see Figure \ref{fig:domainerror}). We find that our coherence fine-tuned (CO) model always performs better than or on par with the supervised baseline (SU) model, expect in the case of Phrase Detectives - Gutenburg (PD-G). We postulate that the increase in PD-G errors can be attributed to the length of the documents in the dataset.\footnote{The average document length of PD-G is 1507.2 tokens, which is the highest among all datasets.}

\begin{figure}[t]
    \centering
    \includegraphics[width=\columnwidth]{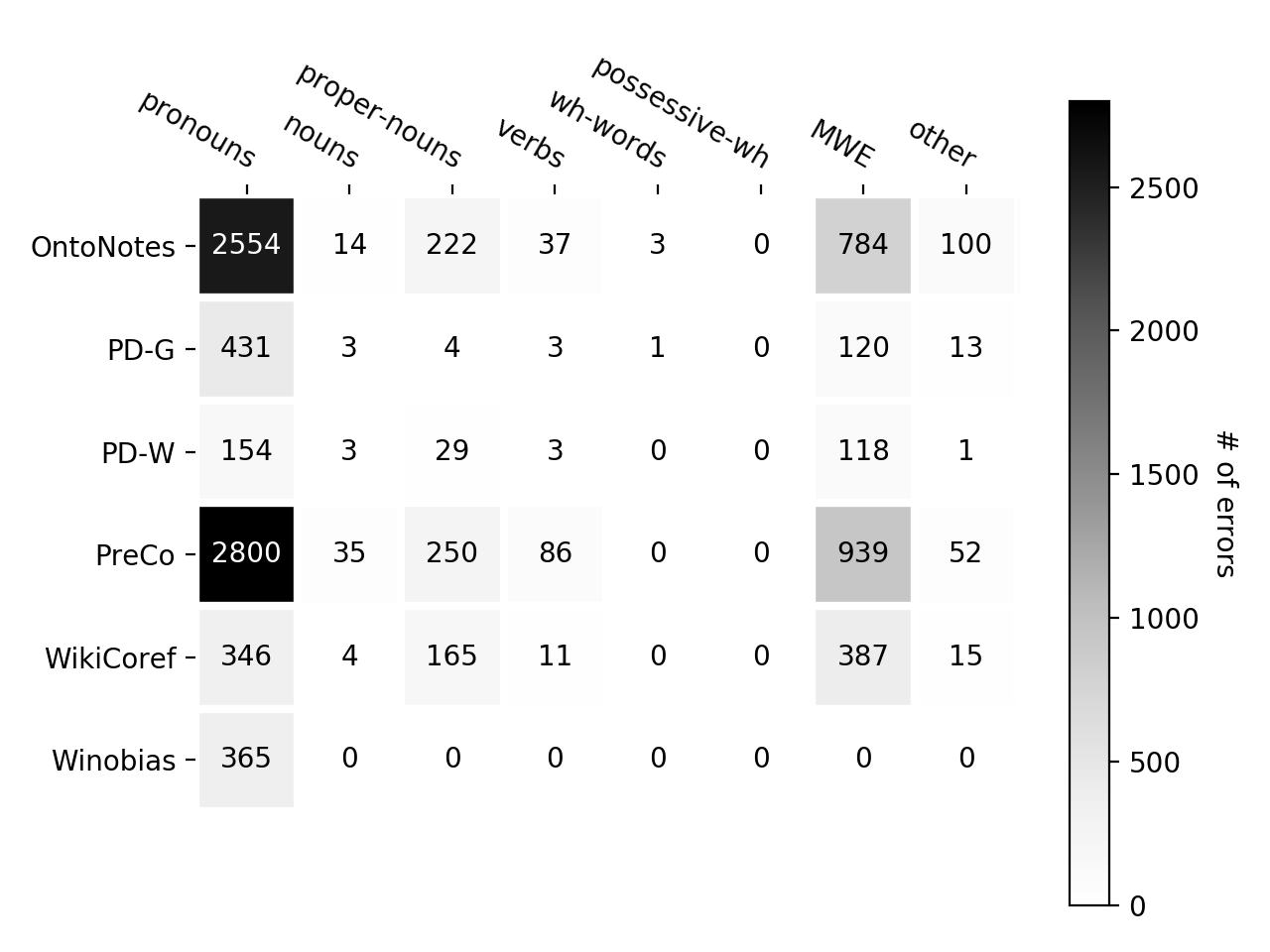}
    \caption{Heatmap showing the POS-tag categories for the antecedents that our fine-tuned coreference system incorrectly classified. All domains except WikiCoref have the highest amount of errors made when the antecedent is a pronoun. Here, pronouns are PRP, PRP\$; MWE is any multi-word expression, nouns are NN, NNS; proper-nouns are NNP, NNPS; verbs are VB, VBD, VBG, VBN, VBP, VBZ; other tags we observed were IN, JJR, JJ, RB, DT, CD, MD, POS; and wh-words are WDT, WRB, WP, WP\$.}
    \label{fig:poserror}
\end{figure}

\paragraph{Part-of-speech}
As seen in Figure~\ref{fig:poserror}, across all domains, most errors from the coherence fine-tuned system occur when the antecedent is a pronoun, except for WikiCoref, where the most errors occur when the antecedent was a multi-word expression. This trend is seen in the supervised baseline models as well. 

Apart from being the most frequent among mentions, two possible reasons why pronouns could be predicted incorrectly most often are: (i) as the distance in text increases between the original antecedent and subsequent pronouns, it becomes more difficult to resolve, and (ii) as a text becomes more complex, with multiple possible antecedents to choose from, linking becomes harder. Given the increased performance of our coreference resolver from the inclusion of a coherence classifier, we hypothesize that the second problem would be easier for our system to overcome, while the first could still persist. 

\paragraph{Span length}
Finally, we analyse the length of the mentions linked by our models. In general, both supervised baseline and coherence fine-tuned models perform similarly for very short (0--3 tokens) and very long (7+ tokens) mentions. However, we see an improvement in linking accuracy of the coherence fine-tuned model when the mention length is between 3--7.


\section{Related Work}\label{sec:related}

\paragraph{Augmented Coreference Resolution}

Previous work has augmented Coreference resolvers with syntax information \cite{wiseman-etal-2016-antecedent,clark-manning-2016-deep,clark-manning-2016-improving}, external world knowledge \cite{rahman2011coreference, emami-etal-2018-generalized,aralikatte-etal-2019-rewarding} and a variety of other linguistic features \cite{ng2007shallow,haghighi2009simple,zhang-etal-2019-incorporating}. Similarly, \citet{ponzetto-strube-2006-exploiting,ponzetto-strube-2006-semantic} used features from SRL and external sources for a non-neural coreference resolver.

\paragraph{Augmented Semantic Role Labelling} SRL systems have long utilised annotations from syntactic formalisms as an essential component \cite{levin1993english, hacioglu2004semantic, pradhan2005semantic, sutton2005joint, punyakanok2008importance}. More recently, \citet{strubell2018linguistically} showed that it was possible to exploit information from syntactic parses for supervision of the self-attention mechanism in a fully differentiable transformer-based SRL model, surpassing the previous state-of-the-art. \citet{xia2019syntax} follow up on this, presenting a detailed investigation into various methods of incorporating syntactic knowledge into neural SRL models, finding it consistently beneficial. 

\paragraph{Document level consistency}

Document-level modelling has been shown to be beneficial for NLP tasks such as machine summarization \cite{chen2016distraction}, translation \cite{maruf-haffari-2018-document,voita-etal-2018-context, junczys2019microsoft}, sentiment analysis \cite{bhatia-etal-2015-better}, and question answering \cite{verberne2007discourse, sadek2016discourse}. For semantic analyzers, document-level consistency is an important requirement. Indeed, when training on complete documents, it also provides a strong input signal. In previous work \citet{tang-2015-consistent} presented a user product neural network and validated the effects of users and products in terms of sentiment and text-based consistency. Likewise, \citet{du-etal-2019-consistent} used label consistency as an additional objective for a procedural text comprehension model, showing state-of-the-art performance. More recently, \citet{liu-lapata-2018-learning} used discourse structure and global consistency to guide a machine comprehension model. 

Our approach is orthogonal and possibly complementary to those described above: we investigate the consistency in the overall information presented in complete documents for span graphs composed of semantic role labeling and coreference resolution annotations.


\section{Conclusion}\label{sec:conclusion}

We presented a joint coreference resolver and semantic role labeler along with a method of fine-tuning them with document-level coherence rewards over unlabeled documents. We find that this leads to considerable performance gains for coreference resolution across domains, and moderate improvements for semantic role labeling. Results are presented across six English coreference resolution datasets and four English semantic role labeling datasets. Our code will be made publicly available at \url{https://github.com/rahular/joint-coref-srl}

Future work will improve the efficiency of our training procedure to allow fine-tuning on longer documents, and investigate how the models can be further improved with better credit assignment.

\appendix
\section{Results}\label{sec:single-task_results}

The single-task results on the six coreference resolution and four SRL datasets are shown in Tables~\ref{tab:coref-single-task-results} and \ref{tab:srl-single-task-results} respectively. The multi-task results are shown in Tables~\ref{tab:coref-multi-task-results} and \ref{tab:srl-multi-task-results} respectively. The last column in each table shows the \fone score averaged across all datasets for a given model.

\begin{table*}[th]
\centering
\scriptsize
\begin{tabular}{@{}r|cccccccccccc|cc@{}}
\toprule
 & \multicolumn{2}{c}{\textbf{OntoNotes}} & \multicolumn{2}{c}{\textbf{PreCo}} & \multicolumn{2}{c}{\textbf{PD-G}} & \multicolumn{2}{c}{\textbf{PD-W}} & \multicolumn{2}{c}{\textbf{WikiCoref}} & \multicolumn{2}{c}{\textbf{WinoBias}} & \multicolumn{2}{|c}{\textbf{Average}} \\ 
\multirow{-2}{*}{} & {\textbf{Baseline}} & {\textbf{Ours}} & {\textbf{Baseline}} & {\textbf{Ours}} & {\textbf{Baseline}} & {\textbf{Ours}} & {\textbf{Baseline}} & {\textbf{Ours}} & {\textbf{Baseline}} & {\textbf{Ours}} & {\textbf{Baseline}} & {\textbf{Ours}} & {\textbf{Baseline}} & {\textbf{Ours}} \\
\midrule
{\textbf{LSTM + CNN}}   & {62.58} & {62.59} & {42.36} & {44.37} & {46.99} & {46.73} & {32.72} & {33.24} & {39.80} & {39.82} & {69.61} & {69.68} & {49.01} & {\bf 49.40} \\
{\textbf{{\sc BERT}-Tiny}}    & {61.01} & {61.35} & {42.26} & {45.41} & {48.44} & {48.93} & {37.65} & {37.76} & {45.16} & {45.03} & {63.68} & {67.23} & {49.70} & {\bf 50.95} \\
{\textbf{{\sc BERT}-Mini}}    & {64.20} & {64.15} & {44.82} & {46.07} & {49.95} & {50.31} & {40.22} & {40.26} & {49.70} & {49.44} & {66.76} & {67.07} & {52.61} & {\bf 52.88} \\
{\textbf{{\sc BERT}-Small}}   & {65.81} & {66.39} & {44.32} & {47.51} & {51.25} & {52.14} & {42.35} & {42.11} & {50.30} & {50.55} & {62.53} & {64.72} & {52.76} & {\bf 53.90} \\
{\textbf{{\sc BERT}-Medium}}  & {68.46} & {68.72} & {45.85} & {48.53} & {54.65} & {54.68} & {42.41} & {41.80} & {54.05} & {54.20} & {68.57} & {69.23} & {55.67} & {\bf 56.19} \\
{\textbf{{\sc BERT}-Base}}    & {71.48} & {71.53} & {46.89} & {48.61} & {56.87} & {57.27} & {44.79} & {44.16} & {55.02} & {55.32} & {71.62} & {72.17} & {57.78} & {\bf 58.18} \\
\bottomrule
\end{tabular}
\caption{{\sc Coreference resolution} results of single-task models. `Baseline' and `Ours' indicate the average \fone scores of \muc, \bcubed and \ceaf for the supervised baseline and coherence fine-tuned models respectively. PD-(G/W) --- Phrase Detectives (Gutenberg/Wikipedia) splits.}
\label{tab:coref-single-task-results}
\end{table*}

\begin{table*}[h]
\centering
\scriptsize
\begin{tabular}{@{}r|cccccccccccc|cc@{}}
\toprule
 & \multicolumn{2}{c}{\textbf{OntoNotes}} & \multicolumn{2}{c}{\textbf{PreCo}} & \multicolumn{2}{c}{\textbf{PD-G}} & \multicolumn{2}{c}{\textbf{PD-W}} & \multicolumn{2}{c}{\textbf{WikiCoref}} & \multicolumn{2}{c}{\textbf{WinoBias}} & \multicolumn{2}{|c}{\textbf{Average}} \\ 
\multirow{-2}{*}{} & {\textbf{Baseline}} & {\textbf{Ours}} & {\textbf{Baseline}} & {\textbf{Ours}} & {\textbf{Baseline}} & {\textbf{Ours}} & {\textbf{Baseline}} & {\textbf{Ours}} & {\textbf{Baseline}} & {\textbf{Ours}} & {\textbf{Baseline}} & {\textbf{Ours}} & {\textbf{Baseline}} & {\textbf{Ours}} \\
\midrule
{\textbf{LSTM + CNN}}   & {62.13} & {62.15} & {42.77} & {47.66} & {47.06} & {47.10} & {35.23} & {35.54} & {40.47} & {41.01} & {64.23} & {64.13} & {48.65} & \textbf{49.60} \\
{\textbf{BERT Tiny}}    & {59.76} & {60.53} & {42.22} & {49.11} & {42.58} & {42.22} & {35.46} & {35.53} & {46.68} & {47.90} & {47.23} & {71.73} & {45.65} & {\bf 51.17} \\
{\textbf{BERT Mini}}    & {63.43} & {63.80} & {44.18} & {46.40} & {47.11} & {47.56} & {38.95} & {39.09} & {51.32} & {51.89} & {55.88} & {69.45} & {50.14} & {\bf 53.02} \\
{\textbf{BERT Small}}   & {65.40} & {65.75} & {44.91} & {46.82} & {51.29} & {51.42} & {41.33} & {40.72} & {51.72} & {52.24} & {52.89} & {65.30} & {51.26} & {\bf 53.73} \\
{\textbf{BERT Medium}}  & {67.70} & {68.06} & {45.99} & {47.52} & {53.65} & {53.21} & {42.65} & {42.80} & {52.94} & {53.30} & {45.97} & {68.46} & {51.48} & {\bf 55.52} \\
{\textbf{BERT Base}}    & {70.78} & {71.23} & {47.29} & {48.23} & {55.46} & {55.32} & {43.80} & {43.50} & {57.78} & {57.53} & {63.29} & {69.62} & {56.40} & {\bf 57.55} \\
\bottomrule
\end{tabular}
\caption{{\sc Coreference resolution} results of multi-task models. `Baseline' and `Ours' indicate the average \fone scores of \muc, \bcubed and \ceaf for the supervised baseline and coherence fine-tuned models respectively. PD-(G/W) --- Phrase Detectives (Gutenberg/Wikipedia) splits.}
\label{tab:coref-multi-task-results}
\end{table*}

\begin{table*}[th]
\centering
\small
\begin{tabular}{@{}r|cccccccc|cc@{}}
\toprule
 & \multicolumn{2}{c}{\textbf{OntoNotes}} & \multicolumn{2}{c}{\textbf{Conll05-WSJ}} & \multicolumn{2}{c}{\textbf{Conll05-Brown}} & \multicolumn{2}{c}{\textbf{EWT}} & \multicolumn{2}{|c}{\textbf{Average}} \\
\multirow{-2}{*}{} & {\textbf{Baseline}} & {\textbf{Ours}} & {\textbf{Baseline}} & {\textbf{Ours}} & {\textbf{Baseline}} & {\textbf{Ours}} & {\textbf{Baseline}} & {\textbf{Ours}} & {\textbf{Baseline}} & {\textbf{Ours}} \\ 
\midrule
{\textbf{LSTM + CNN}}   & {72.14} & {72.14} & {67.33} & {67.72} & {63.41} & {63.26} & {67.64} & {67.84} & {67.63} & \textbf{67.74} \\
{\textbf{{\sc BERT}-Tiny}}    & {65.05} & {65.01} & {53.02} & {52.91} & {51.62} & {52.56} & {57.79} & {57.83} & {56.87} & {\bf 57.08} \\
{\textbf{{\sc BERT}-Mini}}    & {77.32} & {77.34} & {68.74} & {68.59} & {65.75} & {65.46} & {70.22} & {70.53} & {\bf 70.51} & {70.48} \\
{\textbf{{\sc BERT}-Small}}   & {81.21} & {81.21} & {73.11} & {73.45} & {69.80} & {70.42} & {72.91} & {72.85} & {74.26} & {\bf 74.48} \\
{\textbf{{\sc BERT}-Medium}}  & {82.30} & {82.32} & {75.24} & {75.21} & {70.21} & {69.96} & {74.73} & {74.79} & {\bf 75.62} & {75.57} \\
{\textbf{{\sc BERT}-Base}}    & {85.88} & {85.97} & {78.93} & {78.83} & {75.01} & {75.30} & {78.00} & {77.99} & {79.46} & {\bf 79.52} \\
\bottomrule
\end{tabular}
\caption{{\sc Semantic role labeling} results of single-task models. `Baseline' and `Ours' indicate the average token \fone scores of the supervised baseline and coherence fine-tuned models respectively.}
\label{tab:srl-single-task-results}
\end{table*}

\begin{table*}[th]
\centering
\small
\begin{tabular}{@{}r|cccccccc|cc@{}}
\toprule
 & \multicolumn{2}{c}{\textbf{OntoNotes}} & \multicolumn{2}{c}{\textbf{Conll05-WSJ}} & \multicolumn{2}{c}{\textbf{Conll05-Brown}} & \multicolumn{2}{c}{\textbf{EWT}} & \multicolumn{2}{|c}{\textbf{Average}} \\
\multirow{-2}{*}{} & {\textbf{Baseline}} & {\textbf{Ours}} & {\textbf{Baseline}} & {\textbf{Ours}} & {\textbf{Baseline}} & {\textbf{Ours}} & {\textbf{Baseline}} & {\textbf{Ours}} & {\textbf{Baseline}} & {\textbf{Ours}} \\ 
\midrule
{\textbf{LSTM + CNN}}   & {72.92} & {72.58} & {68.02} & {67.82} & {60.98} & {60.52} & {67.21} & {67.27} & {\bf 67.28} & {67.05} \\
{\textbf{BERT Tiny}}    & {63.91} & {64.03} & {52.16} & {52.19} & {52.60} & {52.97} & {57.94} & {58.23} & {56.65} & {\bf 56.85} \\
{\textbf{BERT Mini}}    & {77.68} & {77.69} & {66.10} & {66.19} & {69.83} & {69.86} & {70.78} & {70.77} & {71.10} & {\bf 71.13} \\
{\textbf{BERT Small}}   & {81.08} & {81.10} & {69.89} & {70.07} & {73.45} & {73.67} & {74.48} & {74.25} & {74.72} & {\bf 74.77} \\
{\textbf{BERT Medium}}  & {84.45} & {84.47} & {73.05} & {73.35} & {77.52} & {77.68} & {76.55} & {76.55} & {77.89} & {\bf 78.01} \\
{\textbf{BERT Base}}    & {86.41} & {86.40} & {76.59} & {76.47} & {79.34} & {79.30} & {78.67} & {78.58} & {\bf 80.25} & {80.19} \\
\bottomrule
\end{tabular}
\caption{{\sc Semantic role labeling} results of multi-task models. `Baseline' and `Ours' indicate the average token \fone scores of the supervised baseline and coherence fine-tuned models respectively.}
\label{tab:srl-multi-task-results}
\end{table*}

\bibliography{main}

\end{document}